# Efficient Deep Learning on Multi-Source Private Data

Nick Hynes    Raymond Cheng    Dawn Song

UC Berkeley

{nhynes, ryscheng, dawnsong}@berkeley.edu

*Abstract*—Machine learning models benefit from large and diverse datasets. Using such datasets, however, often requires trusting a centralized data aggregator. For sensitive applications like healthcare and finance this is undesirable as it could compromise patient privacy or divulge trade secrets. Recent advances in secure and privacy-preserving computation, including trusted hardware enclaves and differential privacy, offer a way for mutually distrusting parties to efficiently train a machine learning model without revealing the training data. In this work, we introduce Myelin, a deep learning framework which combines these privacy-preservation primitives, and use it to establish a baseline level of performance for fully private machine learning.

## I. INTRODUCTION

Machine learning (ML) has enabled a variety of applications from smart-homes to personal assistants. Such success is largely due to recent algorithmic breakthroughs, increased availability of computational resources, and access to vast quantities of data which enable training complex models. However, such datasets often contain sensitive information and therefore raise several privacy concerns. For instance, it has recently been demonstrated that personally identifying information can be inferred from even rough estimates of an ad campaign's audience size [29].

Additionally, machine learning models can benefit from multiple providers' shared data. Examples include clinical researchers training a model on patient information from several geographically distributed clinics, and banks pooling data to train a higher quality fraud detection model. In both cases, directly sharing data is unacceptable: the clinics must protect their patients' privacy and banks desire to protect their trade secrets. Thus, protecting the confidentiality of the data, the model, and the computations on them requires a privacy-preserving machine learning platform.

Although it is currently possible to train fully private individual ML models, existing systems for privacy-preserving machine learning are unable to accommodate to the large-scale, highly flexible deep learning models which power modern ML services. For instance, federated learning [26], designed to allow multiple users to jointly train a private model, has been shown to be vulnerable against certain privacy attacks [17]. Approaches based on direct cryptographic operation over models and data incur a performance penalty of 3 to 4 orders of magnitude. Moreover, for those which depend on trusted hardware, supporting deep learning workloads requires including large libraries which were designed with performance–not privacy–as a main objective.

In this work, we introduce Myelin, a system designed for efficient *differentially-private* and *data-oblivious* deep learning in trusted hardware *enclaves*.

**Contributions** We demonstrate the base performance of Myelin through benchmarks on practical ML models. This paper contributes a system for efficient, fully-private training of ML models in hardware enclaves. We demonstrate state-of-the-art single-enclave performance through benchmarks on practical ML models.

Importantly, Myelin can be deployed on existing commodity hardware and offers a tool for both production applications as well as continued exploration of the problem space.

## II. BACKGROUND

### A. Machine Learning

A common goal of machine learning is to find a model $f_\theta(x)$ which minimizes a loss function $\mathcal{L}$ by adjusting parameters $\theta$ based on observations $x$. Models can range in complexity from simple linear models (e.g., $f_\theta(x) = \langle x, \theta \rangle$ ) to deep neural networks composed of linear models and non-linearities. In either case, ensuring privacy of data requires protecting both the data *and* the parameters [4]. Indeed, for vectors $x$ and $\theta$, an algorithm could simply set $\theta = x$.

For the model optimization algorithm, we focus on stochastic gradient descent (SGD), the workhorse of modern, scalable model training [13]. At its most basic, an SGD update takes the form

$$\theta^{(t+1)} = \theta^{(t)} - \eta \nabla_\theta \mathcal{L}(f_{\theta^{(t)}}, x) \qquad (1)$$

where $\nabla_\theta$ is the gradient of the objective function with respect to the parameters and $\eta$ is the step size, or learning rate. Intuitively, SGD finds the direction which maximally increases the loss and takes a small step in the *opposite* direction. SGD is typically performed in a loop which presents a *mini-batch* of $m$ examples to the model and then updates using the average of their gradients to reduce variance.

### B. Trusted Hardware Enclaves

A trusted hardware enclave is a protected environment in which code can run without external observation and generate a report verifying its own identity. This is achieved through memory isolation (e.g., encryption) , a secure random number generator, and a trusted monitor which manages the enclave's interaction with the untrusted world [11]. The threat model is that of a malicious adversary who can manipulate any aspect of the system except the contents of the CPU, itself.

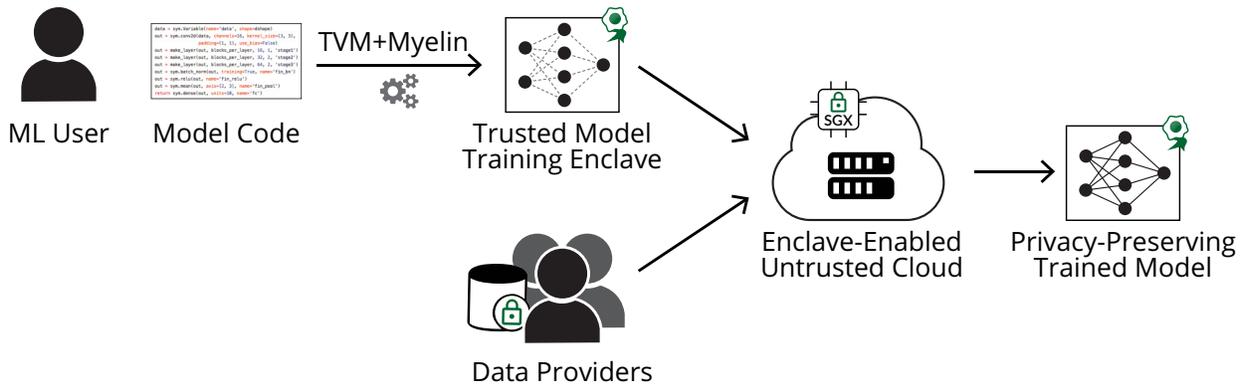

Fig. 1: Overview of Myelin, a system for efficient, distributed, privacy-preserving machine learning. An ML user's model is automatically compiled into a high-performance library by TVM (Tensor Virtual Machine) and made private using Myelin. The model is then privately trained using trusted hardware on privately shared data. The system is described in detail in Section IV.

In addition to having isolated memory, it is necessary than an enclave be able to attest to its own identity once deployed. This is generally enabled by a hardware-based root of trust [25]. Myelin depends on remote attestation to prove its identity and establish secure communication channels with the data consumer and each of the data providers.

In this work, we use enclaves provided by Intel Software Guard eXtensions (SGX) [2]. In SGX, the root of trust is Intel, their private key, and the keys which they assign to each processor. Although trusting Intel is a strong assumption, we note that the techniques introduced to follow generalize to any enclave implementation (e.g., [11]).

### C. Differential Privacy

Applying differential privacy to an algorithm provides a strong bound on how much information it can leak about any item of input data. Formally, for all datasets $D_i, D_j \in D$ $D_i, D_j$ which differ in a single entry, if

$$\Pr[\mathcal{A}(D_i) \in r] \leq e^\epsilon \Pr[\mathcal{A}(D_j) \in r] + \delta \quad \forall r \subseteq R$$

then $\mathcal{A}$ is $(\epsilon, \delta)$-differentially private [10]. This differs from the original definition of $\epsilon$-DP [9] by the addition of the $\delta$ term which allows $\epsilon$-DP to be broken with probability $\delta \ll 1/|D_i|$ where $|D_i|$ is the size of the dataset. Myelin relies on differential privacy to allow data consumers to query the trained model without revealing the training data.

Differential privacy via *output perturbation* is achieved using a mechanism $\mathcal{M}_{\epsilon,\delta} : \mathcal{A} \times \mathcal{D} \to \mathcal{A}'$ which adds noise so that that $\mathcal{A}'$ is $(\epsilon, \delta)$-DP. The amount of the noise depends on the *sensitivity* of $\mathcal{A}$, which is defined as

$$S_\mathcal{A} = \max_{D_i, D_j} \frac{|\mathcal{A}(D_i) - \mathcal{A}(D_j)|}{|D_i - D_j|}$$

Each query to $\mathcal{A}'$ incurs a privacy cost of $\epsilon$, so an algorithm that calls $\mathcal{A}'$ $n$ times is $(n\epsilon, n\delta)$-DP. This result is known as *strong composition* [21].

In the context of machine learning, $\mathcal{A}$ is the learning algorithm which trains the model and returns the model parameters. In Myelin, the learning algorithm is SGD and therefore $S_\mathcal{A}$ is the norm of the gradient : $\max_{D_i} \|\nabla_\theta f_\theta(D_i)\|$ [27]. Thus, modifying SGD to produce a differentially private model requires normalizing the $\ell_2$ norm of the gradient so that it does not exceed a bound $B$ and applying $\mathcal{M}_{\epsilon,\delta}$ to this *clipped* gradient. The noisy gradient is then averaged over the $L$ examples in each *lot* of randomly sampled training instances (a lot can be made of several mini-batches). In this work we use the Gaussian mechanism and the moments accountant [1] which, together, offer tight privacy bounds. The Gaussian mechanism for differentially private SGD is defined as

$$\mathcal{M}_{\epsilon,\delta}(f_\theta, D_i) = \text{clip}(\nabla_\theta f(D_i)) + \mathcal{N}\left(0, \frac{2B^2}{\epsilon^2} \log \frac{5}{4\delta}\right) \quad (2)$$

and the differentially private SGD update is

$$\theta^{(t+1)} = \theta^{(t)} - \eta(\mathcal{M}_{\epsilon,\delta}(f_{\theta^{(t)}}, x)) \quad (3)$$

The moments accountant tracks the log-moments of the privacy loss random variable and provides tighter guarantees than linear strong composition.

### III. PROBLEM DEFINITION & APPROACH OVERVIEW

We formalize our approach using the following notations. Let $P = \{p_1, \ldots, p_n\}$ represent data providers and $C$ be a data consumer who wishes to train a single machine learning model on the data of $P$. The data are sensitive and economically valuable, so provider $p_i$ desires they remain as private as possible while satisfying the utility requirements of consumer $C$. Similarly, $C$ wishes to extract maximal utility from the data—via the model—given the privacy requirements of all providers $P$. Since $C$ has devoted resources to designing the model and procuring the data, $C$ wants exclusive access to the trained model and, therefore, that the model parameters remain private. We consider a scenario in which $P$ and $C$ agree beforehand on the values of privacy and utility.

#### A. Threat Model

For the purposes of this work, we assume that each $p_i$ honestly provides data, so data valuation and poisoning attacks are out of scope. Additionally, we assume that the owner of the enclave-enabled hardware (e.g., cloud computing service operator) will not deny service to the system, so we do not seek to provide liveness guarantees. Otherwise, all parties are actively malicious and wish to learn the contents of the datasets. We also assume that the owner of the trusted hardware—if not the data consumer–would like to obtain the model parameters and use it to perform inference without having to agree to the terms set by the data providers . To these ends, an adversary can observe and modify any runtime aspect of the untrusted software or physically accessible computing hardware. This means that an attacker can read the contents of DRAM, report fake data from syscalls (e.g., `time`, `socket`) , observe interrupts, and arbitrarily schedule program (re-)execution. The model assumes that CPU internals like registers and caches are physically inaccessible and safe from outside observation.

Enclave memory is protected either by encryption or physical inaccessibility. However, enclaves are still vulnerable to side- and, for some implementations, controlled-channel [30] attacks. To establish end-to-end privacy, Myelin mitigates these risks by using data-oblivious [23] algorithms. Additionally, Myelin is designed so that the privacy-preserving aspects are enabled by default and transparent to the end user.

*B. Overview*

Using the notation from Section III, we suppose the following workflow which is depicted in Figure 1. The process begins with each data provider $p_i$ uploading an encrypted dataset to a storage service (e.g., AWS, IPFS) and providing an interface to the decrypted data, $D_i$. For simplicity, one may consider a $D_i$ which yields chunks of data to entities on a whitelist. Some time later, the data consumer $C$ deploys an enclave $E$ which proves its identity (or other attributes) to each $D_i$ using remote attestation. The enclave then begins fetching data from the $D_i$s and training the machine learning model. The trust of $p_i$ in $E$ is based on manual or automatic verification of the algorithm run by $E$; this trust is encoded in $D_i$. Once the algorithm has run to completion, $C$ is allowed to make unlimited offline queries of the trained model.

Given our threat model, a fully private training algorithm must be able to run in a hardware enclave and, additionally, be differentially private and data-oblivious [23].

IV. MYELIN: EFFICIENT FULLY PRIVATE ML

The purpose of Myelin is to facilitate high-performance, privacy-preserving machine learning training and inference on multiple distrusting users' data.

*A. Privacy*

In its default mode of operation, Myelin runs fully within an enclave to provide strong privacy guarantees based on data-oblivious and differentially private algorithms. The enclave, verified through remote attestation, ensures that the privacy-preservation mechanisms are faithfully applied.

The fully private training workflow, shown in Figure 2, begins with the data consumer deploying a Myelin enclave to an enclave-enabled device. The enclave then uses remote attestation to establish a secure communication channel with each of the data provider interfaces. The enclave then randomly initializes all model parameters (as is standard practice in ML) and begins the training loop which 1) requests a fixed-sized chunk of each providers' data and aggregates them into a mini-batch, 2) obtains the model's predictions on the mini-batch, 3) computes gradients and performs a differentially private parameter update. During training, all cleartext data and parameters reside in enclave memory and are therefore protected from observation. Differential privacy then provides an upper bound on the amount of information the parameters leak about any provider's data—even when the data consumer is allowed unrestricted offline black-box access to the model.

Although the enclave protects the contents of the CPU and memory, data may still be exfiltrated through side-channel attacks. For instance, using maliciously constructed inputs, timing the `max` operator can reveal the value of a target input. A similar attack can expose the value of a floating point number by measuring the extra cycles taken when the value is *sub-normal* (very near zero) [3]. To mitigate the risk of side-channel attacks, Myelin implements data-oblivious [23] algorithms, such that disk, memory, and network accesses do not depend on the data.

Network obliviousness is achieved by requesting fixed-size chunks of data from all data provider interfaces ($D_i$) at each training iteration; training does not proceed until all providers have yielded a chunk of data. For protecting the computation on received data, we build on the data-oblivious primitives introduced by Ohrimenko et al. and extend them to provide protection of model parameters against sub-normal floating-point timing attacks. Since the dense matrix operations used in deep learning models are already data-oblivious, it mostly suffices for Myelin to replace data-dependent operators (e.g., max-pooling, one-hot encoding) with oblivious variants as in [23]. To prevent leaking information about sub-normal values, the data and parameters are perturbed with a small-but-not-subnormal random number (e.g. $1^{-10}$); this is already done for the weights by the DP mechanism. Model accuracy is not degraded by either modification–indeed, it is possible to train accurate models with even 16-bit fixed-point numbers [15]. Computational overhead is kept to a minimum by custom ML operators included in Myelin.

*B. Efficiency*

The efficiency of Myelin is based on the use of modular, optimized numerical libraries for machine learning computations. These libraries are automatically generated using the recently introduced TVM stack [6], which allows fine-grained definition of both primitive operators (e.g., `conv2d`) and whole computation graphs. Thus, instead of bundling a monolithic library like OpenBLAS or LAPACK, Myelin links to a minimal TVM-generated library which includes only the operations needed to train the model. The benefits of doing so are threefold: the generated operators are made faster and more memory efficient by *fusing* consecutive computations, the resulting Myelin binaries become small enough to fit completely within enclave memory, and the TCB is reduced by several orders of magnitude and becomes amenable to manual or even automatic verification. Whereas the naive approach of running a monolithic numerical library like OpenBLAS in a libOS would require a TCB of over 100k lines of code, the Myelin runtime requires only 1.5k lines of enclave-aware Rust code.

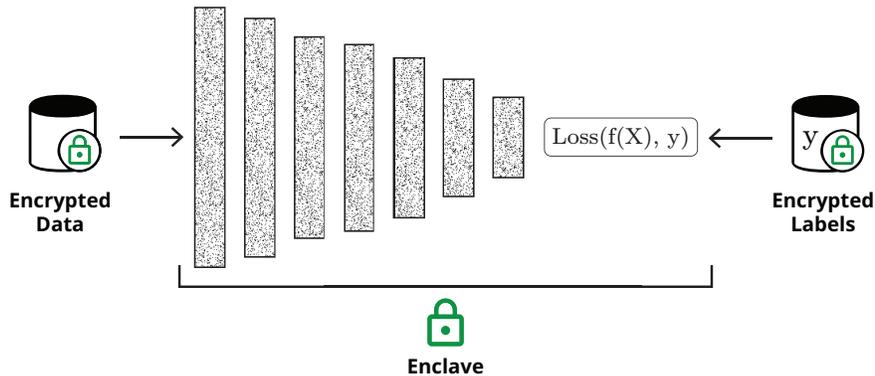

Fig. 2: Efficient data-oblivious and differentially private training within a trusted hardware enclave. All layers' parameters are updated using a DP mechanism (hence the "noisy" background).

*a) Operator Scheduling:* Myelin takes significant advantage of the fine-grained scheduling allowed by TVM to increase both performance *and* privacy. For performance, the single largest improvement is obtained by parallelizing the outer loops of operators like matrix multiplication across several worker threads[1]. Since a single Myelin enclave can fully utilize the CPU of a single machine, the system can make efficient use of a computing cluster through parallel, distributed training. Additionally, Myelin introduces additional fusion rules to hide the latency and allocations introduced by differential privacy. For instance, computing the sensitivity of the training algorithm requires computing the gradients for each example. In common autograd frameworks, calculating $m$ per-example gradients requires copying each parameter $m$ times and applying the forward pass of the model to each example individually (though only requiring a single backward pass) [14]. Notably, this technique does not work for the spatial convolution operator found in virtually all modern computer vision models. In Myelin, through the use of custom TVM operators and schedules, per-example gradients can be obtained simply, and in a single forward and backward pass, by scheduling the sum over the batch dimension as a separate step following gradient clipping and the addition of noise. Moreover, fusing the clip, noise, and sum operations can hide extra latency and allocations required when applying DP.

## V. EXPERIMENTS

To ascertain the performance/utility trade-offs of privacy-preserving machine learning, we benchmark Myelin on the axes of speed and accuracy.

### A. Experiment Setup

The model training benchmarks are performed on the CIFAR-10 image recognition dataset which contains 50k 32x32 color images and is challenging enough to require models that approximate those used on real-world tasks. For baselines, we report performance for the model used in Gazelle [20] and the VGG9 used in Chiron [19]. We also provide results for a very deep-yet-narrow ResNet-32 using the setup described in [16]. Since deeper variants of ResNet have been used to achieve state-of-the-art performance on the large-scale ImageNet classification task, the ability to train shallower ones paves the way to achieving such results *privately*.

We also present results for inference-only Myelin using the MobileNet [18] model trained on ImageNet. Inference in MobileNet takes a full-color 224x224 image and produces predictions over 1000 object categories. Although Myelin does not yet support training the depthwise convolutions used by MobileNet, fully-private inference allows users to make predictions on their own private data.

All experiments use an Intel Xeon E5-2690 v4 CPU.

For differential privacy, parameters are set following the example of [1]: privacy parameter $\epsilon = 4$, DP failure probability $\delta = 10^{-5}$, and lot size $L = 1024$. The noise level is chosen to distribute the privacy budget evenly across all epochs. Since the expenditure of $\epsilon$ places a premium on the number of epochs, we compress training of both ResNet-32 and VGG9 to 75 and 40 epochs, respectively. This was found not to harm performance and, in fact, this modification was motivated by observing that the model had long plateaus in validation accuracy.

---

[1]In Myelin, the TVM operators request threads from the Myelin trusted runtime, which in turn requests threads from the Myelin untrusted runtime. The computation does not proceed until all threads re-enter the enclave and thread allocation is not data dependent, so safety is never violated.

*B. Myelin Benchmarks*

The fully private benchmarks use Myelin in its default mode, which is described in Section IV. We report results for training VGG9 and ResNet-32 using Myelin both with and without privacy preservation techniques; and inference in MobileNet. As a baselines for training speed, we use the recently introduced Chiron framework [19], which is based on trusted hardware enclaves. For inference speed, we compare to Gazelle [20], which provides privacy via cryptographic methods, and Slalom [28], which uses trusted hardware in tandem with an untrusted GPU. Since none of these systems include differential privacy, we compare using non-DP Myelin models.

*C. Results*

| Model | Framework | Speed Train (min/epoch) | Speed Test (img/s) | Test Acc. |
|---|---|---|---|---|
| [22] | Gazelle (HE+GC) | – | 0.08 | 93.1 |
|  | Myelin | 2.28 | 1111 | 93.1 |
| VGG9 | Chiron (4 enclaves) | 6.74 | – | 88.1 |
|  | Myelin (1 enclave) | 6.68 | 521 | 89.5 |
| ResNet-32 | Myelin | 11.4 | 476 | 92.4 |
| MobileNet | Slalom (1 enclave+GPU) | – | 35.7 | 71.0 |
|  | Myelin (1 enclave) | – | 35.1 | 71.0 |

TABLE I: Performance comparison of Myelin with related works. To fairly compare accuracy, the Myelinated models are trained without differential privacy. An *epoch* refers to a complete pass over the training/test set. The CIFAR-10 training set of contains 50k images.

| Model | Training Method | Speed (mins/epoch) | Test Acc. |
|---|---|---|---|
| VGG9 | non-private CPU (baseline) | 6.12 | 89.5 |
|  | Myelin | 6.68 | 84.4 |
| ResNet-32 | non-private CPU (baseline) | 12.3 | 92.4 |
|  | Myelin | 12.9 | 90.8 |

TABLE II: Performance comparison of fully private training for VGG9 and ResNet-32 models on the CIFAR-10 image recognition benchmark. An *epoch* means one complete pass over the dataset.

*1) Fully Private Myelin:* The results of non-private CPU and fully private in-enclave training are shown in Table II. Relative to the non-private CPU scenario, we find that the use of differential privacy and data-oblivious algorithms do not significantly reduce training speed. This is to be expected as the majority of computation involves tight, cache-friendly loops, which avoids the main overheads of context switches and memory encryption. Additionally, the accuracy of the DP models is also comparable to that of the non-private models; this is largely a consequence of the tight privacy loss bounds provided by the Moments Accountant.

Table I shows the comparison to related work. We observe that hardware enclaves offer significantly greater efficiency than cryptographic methods. Furthermore, given the use of enclaves, we observe that a *single* Myelin enclave offers better speed and accuracy than *four* Chiron enclaves running asynchronously in parallel. The slight reduction in accuracy for Chiron is actually due to the asynchrony: its enclaves occasionally produce gradients using stale copies of the model. Indeed, if four Myelin enclaves were deployed on four machines, one would expect a similar reduction in accuracy but with nearly a four-fold reduction in training time. Perhaps more surprising is that for trusted inference, Myelin performs as well as Slalom, which leverages the GPU: Myelin's multi-threaded enclaves avoid the overhead of CPU-GPU memory transfer.

Overall, these results show that privacy-preserving training in hardware enclaves is on par with non-private CPU-based training. Although a single CPU is not as efficient as a GPU, the mass availability of CPUs permits highly parallel distributed training workflows which are heavily used in practice [8].

## VI. RELATED WORK

Related work in privacy-preserving machine learning generally falls into two categories based on the mechanism by which privacy is provided: namely, trusted hardware enclaves and direct cryptographic methods like homomorphic encryption and multi-party computation.

*a) Trusted Hardware Enclaves:* Despite the demonstrated utility of enclaves for data analysis [5], directly using them for machine learning workloads is not wholly straightforward. Primary challenges are 1) enabling training of efficient, flexible models within the enclave, 2) protecting against side- and controlled-channel attacks, and 3) managing data size. An approach to these challenges depends significantly on the intended application.

The work of Ohrimenko et al. [23] is based on a setup similar to ours in that several data providers train a shared ML model. Their focus, however, is on hiding memory access patterns for SGX-only training and does not consider DP.

Since their focus is on mitigating side-channel attacks, they do not directly address the issues of model performance or data size. Myelin builds on this work by providing optimized libraries for training data-oblivious deep learning models on incrementally fetched data.

Concurrent work of Hunt et al. [19] introduces Chiron, a system for privacy-preserving *ML as a service* (MLaaS). MLaaS assumes a single data provider/consumer who trains a model using the algorithms and trusted hardware of an untrusted entity. While Chiron does not support GPU accelerated training and is single threaded within the enclave, it increases throughput by distributing training across several enclaves (though with a slight drop in model accuracy). Additionally, Chiron enclaves must include a model compiler and interpreter, which increases the TCB and adds runtime overhead. Practically, this means that they cannot run on existing commodity hardware. Though the data and model are kept private within the enclave, the outputs are not differentially private and the computation is not oblivious, as in Myelin. Of course, Myelin must take extra precaution to ensure that the multiple data providers cannot learn each others' data during training *or* inference. Efficiency, in our case, is derived from multi-threaded enclaves and privacy-preserving training on accelerated hardware. The modular nature of Myelin enclaves makes distributed training (potentially as part of a larger model) relatively straightforward, though we do not benchmark this feature in this work.

Concurrent work of Tramer and Boneh [28] introduces Slalom, which provides efficient privacy-preserving neural network inference using trusted hardware which delegates matrix multiplication to an untrusted GPU. Although their technique offers greater performance than a single-threaded enclave, the speedup is limited by the requirement that non-linear activation functions be computed within the enclave. While, the cryptographic blinding method used by Slalom does not immediately lend itself to model training, we hypothesize that their inference performance could be improved by incorporating techniques used in Myelin.

*b) Cryptographic Methods:* In the absence of trusted hardware enclaves, primary approaches to privacy-preserving computation include secure multi-party computation (MPC) like Garbled Circuits [31] and homomorphic encryption (HE) [24]. The goal of homomorphic encryption is to compute a function directly on encrypted data. The two-party case of MPC, which finds use in machine learning, constructs a randomized (*garbled*), encrypted Boolean circuit which allows both parties to compute the function encoded in the circuit without revealing their private inputs. Each has its respective benefits and, indeed, can be used in tandem to achieve substantial performance improvements [20]. The benefits of such cryptographic methods are that the root of trust is in the algorithms, themselves, and that they may be used directly with hardware accelerators [12, 7]. Notwithstanding, both HE and MPC come at the cost of three orders of magnitude greater runtime overhead as compared to enclaves.

## VII. CONCLUSION & FUTURE WORK

In this work we evaluated, though Myelin, the performance implications of fully private machine learning on multi-source private data using trusted hardware. In general, we observe that private training is on par with CPU-based training. Additionally, we explored the use of plausible deniability as a method for training a deep learning model on an untrusted GPU. For CPU-GPU training, we find that a large speedup is possible but comes at the price of model accuracy. In either case, by allowing collaboration between mutually distrusting data providers and machine learners, we hope to democratize the development and use of ML-enabled services. To this end, future work includes integrating Myelin into a distributed data and computation marketplace so to enable uncoordinated, autonomous execution of ML models on economically valuable, private data.